\documentclass[runningheads]{llncs}
\usepackage{graphicx}
\usepackage{amsmath}
\usepackage[switch]{lineno}
\usepackage{xcolor}
\usepackage{diagbox}
\usepackage{color}
\usepackage{subfigure}
\usepackage{amssymb}
\usepackage[ruled,vlined]{algorithm2e}
\begin{document}
\title{KnowSR: Knowledge Sharing among Homogeneous Agents in Multi-agent Reinforcement Learning}
\author{Zijian Gao, Kele Xu, Bo Ding, Huaimin Wang, Yiying Li, Hongda Jia}
\authorrunning{Zijian Gao et al.}
\institute{College of Computer,National University of Defense Technology
\email{gaozijian19@nudt.edu.cn, kelele.xu@gmail.com, dingbo@aliyun.com, \{hmwang, liyiying10, jiahongda17\}@nudt.edu.cn}}
\titlerunning{KnowSR: Knowledge Sharing among Homogeneous Agents in MARL}
\maketitle              
%
\begin{abstract}
Recently, deep reinforcement learning (RL) algorithms have made great progress in multi-agent domain. However, due to characteristics of RL, training for complex tasks would be resource-intensive and time-consuming. To meet this challenge, mutual learning strategy between homogeneous agents is essential, which is under-explored in previous studies, because most existing methods do not consider to use the knowledge of agent models. In this paper, we present an adaptation method of the majority of multi-agent reinforcement learning (MARL) algorithms called “KnowSR” which takes advantage of the differences in learning between agents. We employ the idea of knowledge distillation (KD) to share knowledge among agents to shorten the training phase. To empirically demonstrate the robustness and effectiveness of KnowSR, we performed extensive experiments on state-of-the-art MARL algorithms in collaborative and competitive scenarios. The results demonstrate that KnowSR outperforms recently reported methodologies, emphasizing the importance of the proposed knowledge sharing for MARL.
\keywords{Multi-agent Reinforcement Learning  \and Knowledge Sharing \and Knowledge Distillation.}
\end{abstract}
\section{Introduction}
Reinforcement learning (RL) has recently made significant progress in solving complex tasks, including Go \cite{silver2016mastering}, Atari games \cite{mnih2015human}, continuous control tasks \cite{lillicrap2015continuous}. While existing RL methods mainly work on single agent domains, there are a number of real-world applications that include multiple agents like multi-player games \cite{peng2017multiagent} and multi-robot control \cite{matignon2012coordinated}. In order to train agents successfully and effectively in multi-agent environments, agents should not only considering the effects of dynamic environment, but also the influence of the other agents.

Unfortunately, traditional RL methods such as Q-Learning \cite{watkins1992q} or policy gradient \cite{silver2014deterministic} fail to suit multi-agent environments because any single agent have no head for considering non-stationary policies of other agents. To tackle the challenges in multi-agent domains, several approaches of MARL have been proposed. For example, MADDPG \cite{lowe2017multi} firstly overcomes the problem of instability using all agents' state and action as the input of current agent and a centralized critic for all agents. In addition, MAAC \cite{iqbal2019actor} extends prior methods with the idea of attention mechanism to help each agent to select which agents to attend to. There are also some methods working on credit assignment in cooperative settings with global rewards such as \cite{rashid2018qmix} and \cite{foerster2018counterfactual}.

However, most existing methods do not consider the use of other agents' models in the multi-agent environment to assist the current agent's training. They mainly devoted to help agents communicate and cooperate with each other at the algorithmic level. For instance, \cite{jiang2018learning} proposed to divide communication groups with a bi-directional LSTM unit to guide agents for coordinated strategies. The source of knowledge used to train agents is limited to the feedback from environments, which may may limit the agents' learning. 

In this paper, we propose an adaptation method for knowledge sharing called KnowSR, which can be easily deployed in MARL algorithms. The main idea is to provide more useful information for agents though mutual learning between homogeneous agents. 
Our motivation is derived from the fact that, in many real-world environments, it is beneficial for agents to learn from others. For example, in group works, different people may learn at different paces, but they can learn from each other's advice to speed up the learning process. In order to share knowledge between agents, we employ the idea of knowledge distillation (KD) in our work. As we all know, KD \cite{hinton2015distilling} has made great success in the field of computer vision. Here, we assume that the representation of knowledge is the action taken by the agent, and the action is determined by the output of the network. Therefore, for the current agent, the outputs of other agents could be regarded as a potentially advice based on same observations. During the training phase, agents not only aim to achieve higher rewards in the environment but also asked to learn from the advice from others. In this way, agents can learn from varying rewards and also acquire knowledge by the advice from others. In our opinion, the advice from others is also a feasible way to provide more useful informations to current agent. Empirical experiments were conducted to validate the effectiveness of KnowSR. Figure 1 illustrates an example, in which agents must work together to reach the closest target as soon as possible. Agents can not only interacts with the environment through actions but also get advice on how to move based on the same observations from the other agents and share knowledge by advice.

\begin{figure}[h]
  \centering
  \includegraphics[width=\textwidth]{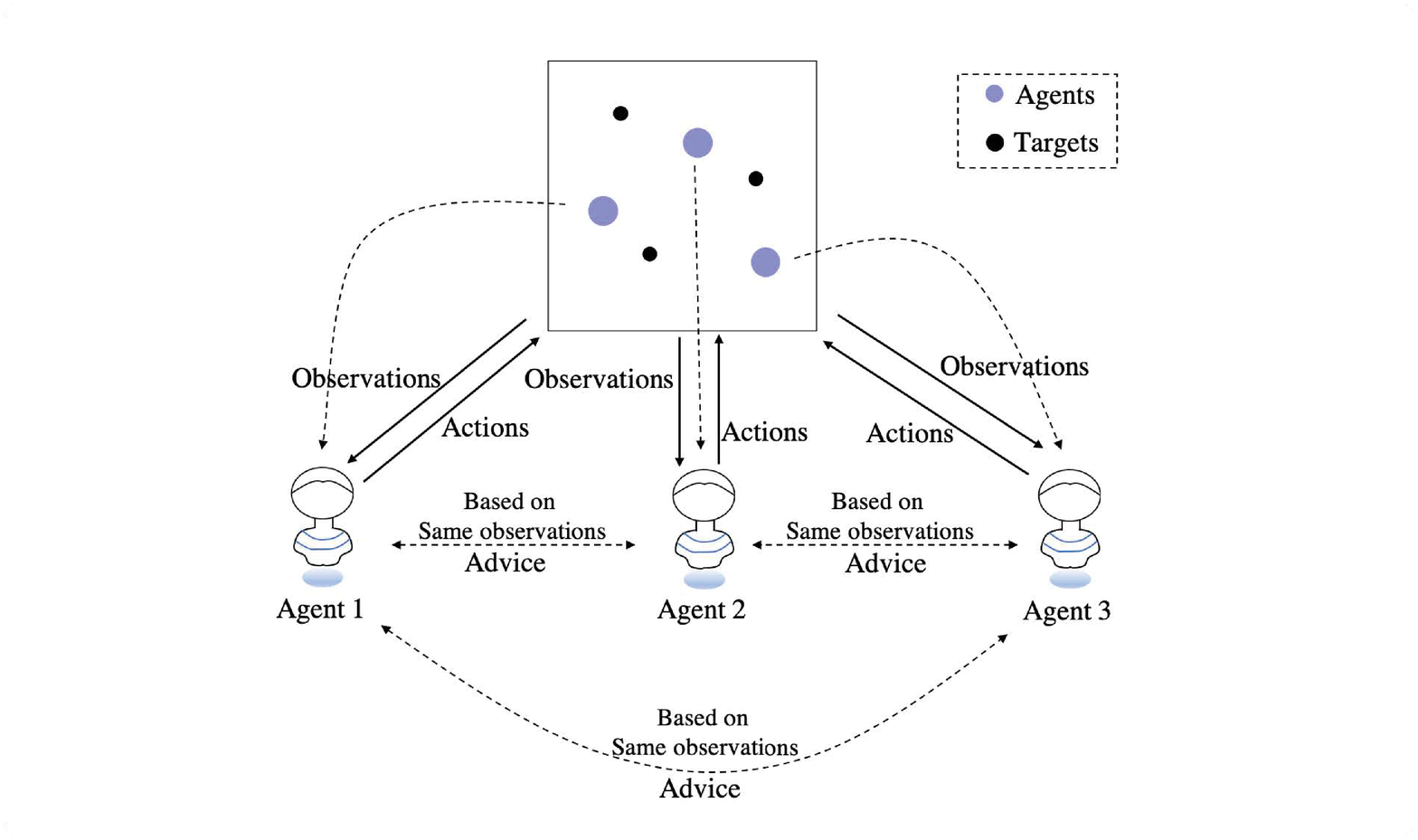}
  \caption{Overview of KnowSR. Agents get advised actions from others in the same situation, and then effectively share the knowledge to help each other learn about what the good strategy is supposed to be.}
  \label{fig:example}
\end{figure}

The contributions in this paper are as follows:
\begin{itemize}
	\item We propose an adaptation framework for MARL, which focuses on accelerating the training phase by knowledge sharing. To the best of our knowledge, it's the under-explored to share knowledge during the training phrase between agents in the previous studies of MARL.
	\item Different strategies were explored to further improve the knowledge sharing performance.
	\item Extensive empirical experiments demonstrate the effectiveness of KnowSR in different task scenarios.
\end{itemize}

\section{Related Work}
Markov Decision Processes (MDPs) \cite{puterman2014markov} is a famous model for sequential decision-making problems, which is used to describe single-agent reinforcement learning system. A system with the Markov property means the state of the next moment is only relevant to the present moment and not related to the previous moments. 
Multi-agent reinforcement learning (MARL) system also has such property, however, compared to the single-agent system, there are multiple agents which are connected with each other under cooperative or competitive relationships. Thus, the traditional independently learning methods does not perform well in practice because they have no head for considering environmental dynamics due to non-stationary policies of other agents such as Q-learning \cite{watkins1992q} and Actor-Critic \cite{konda2000actor}. 

For this reason, considering about some limitations in MARL like nonstationarity of the environment and lack of communication in cooperation settings, there are several methods and their variants have been proposed. As we mentioned, MADDPG, an actor-critic framework which is composed of centralized training with decentralized execution was proposed in \cite{lowe2017multi}. Furthermore, \cite{iqbal2019actor} combined the attention mechanism with centralized critic to help agents focus on vital information. There are also some works focusing on credit assignment. For instance, \cite{foerster2018counterfactual} introduced a centralized critic with shared rewards and allow complex credit assignment though using a “counterfactual baseline”. To conclude, in recent years, the main improving approaches in MARL mostly consider about how to introduce some mechanism to model multi-agent scenarios and relationships. Different from them, in this study, KnowSR put the agent models into consider and make use of the differences in knowledge between models to accelerate training. In comparison, the improvement method KnowSR shows wide application prospects and can be aggregated into MARL algorithms easily.

KD was first proposed in \cite{bucilua2006model} and has since become popular in CV \cite{hinton2015distilling} which is one kind of knowledge transfer method. At beginning of the research, KD was mainly used to compress the knowledge of complex models (\textit{teachers})  into easy and efficient models (\textit{students}) in order to get easy deployable models. Since then, KD also inspires some studies \cite{lai2020dual} \cite{wadhwania2019policy} in RL which are used to compress models, accelerate learning and improve the performances. The main idea of KD is distilling the outputs of teachers and students and calculating the difference between them for backpropagation. The reason why KD distills the outputs is to extract the \textit{dark knowledge}. In our study, our primary focus was how can we get use the idea of KD to solve the problem of agents' slow training in MARL.
\section{Methodology}
\subsection{Preliminaries and Notations}
\textbf {Markov Games}:
A multi-agent extension of Markov Decision Processes (MDPs) \cite{puterman2014markov} can be regarded as partially observable Markov games \cite{littman1994markov}. For N agents in the environment ,they can be defined as tuple, $<S, U, T, R_{1...n},\gamma>$, where $S$ is the state space, $U$ is a collection of action space, $T$ is the state transition function, $A_1...A_n$, $R_{i}$ denotes the reward function of agent $i$, $\gamma$ represents the discount factor. Each agent $i$ has an observation $O_{i}$ in the current state $S_{i}$;
then, each agent use a policy ${\pi}_{\theta_{i}}$ to choose actions and produces the next state according to $T$. Agent $i$ will obtain rewards $R_{i}$ from the environment according to state $S_{i}$ and agent $A_{i}$. The goal of each agent is to maximize its expected discounted returns $R=\sum_{t=0}^{T} \gamma^{t} R_{i}^{t}$, where $t$ is the time horizon.

\textbf {Policy Gradient (PG):} Policy Gradient method \cite{sutton1999policy} aims at maximizing the expected return in few steps with the policy function:

\begin{equation}\label{eq:policy}
\nabla_{\theta} J\left(\pi_{\theta}\right)=\nabla_{\theta} \log \left(\pi_{\theta}\left(a_{t} \mid s_{t}\right)\right) \sum_{t^{\prime}=t}^{\infty} \gamma^{t^{\prime}-t} r_{t^{\prime}}\left(s_{t^{\prime}}, a_{t^{\prime}}\right).\end{equation}

\textbf {Actor-Critic (AC):} In order to ameliorate the high variance of the returns, Actor-critic algorithm \cite{konda2000actor} combines Q-learning and PG and use a value function as the critic to evaluate actions. The critic function can estimate the returns:
\begin{equation}
\label{eq:ACQ}
	Q^{\pi}\left(s_{t}, a_{t}\right)=
	\mathbb{E}_{r_{t}, s_{t+1} \sim E}\left[{r\left(s_{t}, a_{t}\right)+}\right. \\
	 \left. {\gamma \mathbb{E}_{a_{t+1} \sim \pi}\left[Q^{\pi}\left(s_{t+1}, a_{t+1}\right)\right]}\right].
\end{equation}

\textbf {Multi-Agent Deep Deterministic Policy Gradient (MADDPG):} MADDPG \cite{lowe2017multi} which extends from Deep Deterministic Policy Gradient (DDPG) \cite{lillicrap2015continuous} is a successful MARL method and deals with the failure of the not-so-stable performance in multi-agent systems. Because the critic in vanilla DDPG considers only about the local information of the current agent, rather than the global view, DDPG fail to perform well in multi-agent systems. For this reason, MADDPG use a centralized action-value function which takes global observations and actions for all agents as the input: 
\begin{equation}\label{eq:MADDPG}
Q_{i}^{\pi}\left(\mathbf{x}, a_{1}, \ldots, a_{N}\right), \end{equation}
where $\mathbf{x}=\left(o_{1}, \ldots, o_{N}\right)$, $i$ denotes the current agent and let $\pi=\left\{\pi_{1}, \ldots, \pi_{N}\right\}$ be the set of all agent policies. The reason why MADDPG works is that If the all states and actions for all agents are accessible, the multi-agent environment could be regarded to be stable.

\subsection{Knowledge Distillation}
As mentioned earlier, KD has achieved great success in the field of computer vision though extracting \textit{dark knowledge}. The key to extracting \textit{dark knowledge} is to get the soft probability output of trained teachers with distillation. Here, we let $a_{t}$ be the input log of the final softmax layer of the teacher network, where $a_{t}= [a_{1},a_{2},....,a_{j}]$. The logits are converted into probabilities $q_{t}= [q_{1},q_{2},....,q_{j}]$ using the following softmax function: $q_{i}=\frac{e^{a_{i}}}{{\Sigma}_{j} e^{a_{j}}}$ where $i$ represents the $i$-th neuron. \cite{hinton2015distilling} proposed softening the teacher probabilities with temperature $T$ to extract more knowledge:
\begin{equation}\label{eq:soft}
q_{i}=\frac{exp(a_{i}/T)}{{\Sigma}_{j} exp({a_{j}/T})}.
\end{equation}

Compared to true labels from datasets, the soft output of the teacher provides more information. Based on the same input $x$, the teacher and student networks produce probabilities $q_{t}(x)$ and $q_{s}(x)$ with Equation~(\ref{eq:soft}) respectively. The gap between $q_{t}(x)$ and $q_{s}(x)$ is usually penalised by the Kullback--Leibler (KL) divergence (Equation~(\ref{eq:K})):
\begin{eqnarray}\label{eq:KL}
\mathcal{L}_{K D}=T^{2} K L\left(q_{s}(x),q_{t}(x)\right).
\end{eqnarray}
\begin{equation}\label{eq:K}
KL(P \| Q)=\sum P(x) \log \frac{P(x)}{Q(x)}
\end{equation}
where P(x) and Q(x) are two probability distributions of the random variable $x$. KD inspires us to believe that minimizing the gap between agent models through the skill of distillation is the essence of knowledge sharing. We then draw upon the KD thought in our MARL research to share knowledge and verify the feasibility of KnowSR in Section 3.3.

\subsection{Knowledge Sharing via Advice}
Consider a real-world scenario in which the teammates in a group need to cooperate with each other and in the process they need not only to learn by themselves but also to communicate with each other. So, for agents it is also vital to get use of informations from others which can relieve the burden of the high cost of time and resources. We believe that the knowledge that homogeneous agents need to learn is the same, and sharing knowledge is necessary. Thus, mutual learning strategy should be employed in multi-agent systems. However, we have argued that some existing works did not get well used of model-level knowledge. For this reason, it is necessary to find a way of sharing knowledge, in this paper, we aimed at designing a method that works well in such scenarios.

In this study, based on the concept of KD, we propose to share knowledge through advice to minimize the gap between agents in MAS. We designed a task-independent knowledge sharing method that can be applied based on multiple MARL algorithms. First, we make use of the policy models of homogeneous agents. Then, during the training process, the same observations of the current agent were input into both current agent and other homogeneous agents. Based on the same input, the logits, which are the inputs of the last softmax layer, are used to measure the gap between agents with a loss function and we regard the outputs of other agents as advice.  Simultaneously, agents also learn from the environment through returns. 

\begin{figure}[h]
  \centering
  \includegraphics[width=\textwidth]{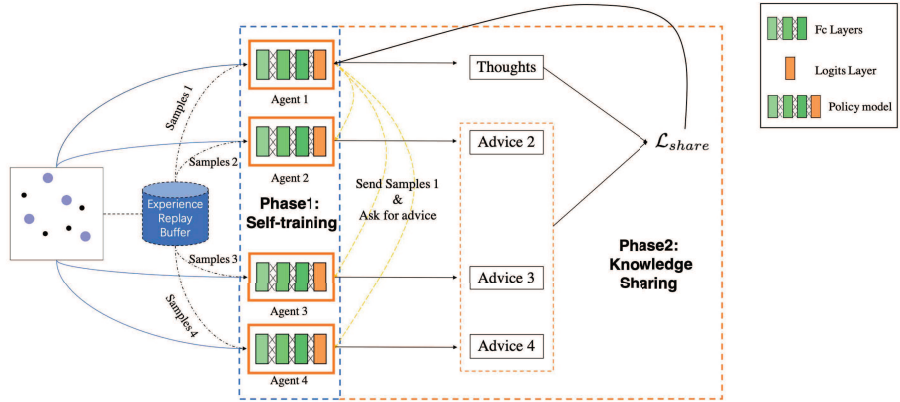}
  \caption{Workflow of KnowSR. 1. Agents learn from experience replay buffer and train by themselves. 2. Each agent send the samples and ask others for advice. Then, calculate the loss based on thoughts and advice to share knowledge. 3. The agent acquires knowledge alternately from environment and other agents.}
  \label{method}
\end{figure}

In this way, the agents are trained by not only maximizing the total return from the environment but also by advice of other homogeneous policy models. As we all know, good advice can be supposed only by competent agents. Agents have to learn knowledge from self-training, which is the foundation of good advice. Hence, self-training and mutual learning must be carried out alternately to ensure the effectiveness of training. Especially in the early training phase, each agent can only explore a subset of the state space, mutual learning can do great help to put the knowledge together and avoid redundant training. Figure~\ref{method} illustrates the main components of KnowSR. KnowSR has been proven to be feasible in various experimental scenarios in Section 4.

\textbf{Advice.}
In the field of computer vision, to extract dark knowledge, it is necessary to soften the teacher model's output probability using Equation~(\ref{eq:soft}) and minimize the gap between them with KL loss using Equation~(\ref{eq:KL}). However, in the MARL, the other agents' policy models also need to learn more knowledge and not authoritative. Therefore, we do not need to soften the probability of the policy models with hyperparameter $T$. Assume that $a_{c}$ and $a_{o}$ are the input logits of the final softmax layer in the current policy model and other agent' policy model, where $a_{c}= [a_{1},a_{2},....,a_{n}]$ and $a_{o}= [a'_{1},a'_{2},....,a'_{n}]$, respectively. Here, we use the mean-squared error (MSE) loss as the $\mathcal{L}_{share}$ loss function to calculate the gap between agents.
\begin{equation}\label{eq:MSE}
\mathcal{L}_{MSE}=\frac{1}{n} \sum_{i=1}^{n}\left({a}_{i}-a'_{i}\right)^{2}.
\end{equation}

\textbf{Self-training and knowledge sharing.}
We state that self-training and sharing knowledge are both important in MARL. Only sufficient self-training can ensure the effectiveness of knowledge sharing, and conversely, knowledge sharing can help agents better self-training. As shown in Figure 2, the training process is divided into two phases: $phase1$ \textbf{Self-training} and $phase2$ \textbf{Knowledge sharing}. In $phase1$, agents learn from experience replay buffer and accumulate knowledge by themselves. As the training progresses, when the agents acquire some knowledge, they enter into the stage of \textbf{Knowledge sharing} stepwise. The output of policy models determines what action should be taken, so, the output of current agent can be regarded as thoughts and the output of other agents can be regarded as advice. At this stage, agent mainly learns from others by comparing the thoughts of current agent and other agents' advice. In the two alternating phases, the agent respectively learns from the reward function based on algorithms and from the advice of other agents. Thus, agents can share knowledge with each other to narrow the space for exploration and accelerate training. Algorithm 1 depicts the KnowSR algorithm based on AC framework.
\begin{algorithm}[htb]
\caption{The training process based on AC framework}

\textbf{Initialization:} Parameters $\theta_{A}$ and $\theta_{C}$ represent current agent's actor and critic networks, respectively; parameter $\theta_{k}$ represents agent k's policy model; $T$ denotes the frequency of update; $I$ denotes the number of agents.

\textbf{Output:} Parameter $\theta_{A}$ of the actor networks and parameter $\theta_{C}$ of the critic networks
for every agent:

counter $M=0$

\textbf{for} episode=1 to max-episodes:

\quad\textbf{for} step $i$=1 to max-steps-in-episode:

\quad\quad take action $a_i$ : $a_i=\pi_{\theta_{A}}(s_i)+\mathcal{N}_i$

\quad\quad get $r_i$ from environment, and observes new state $s_{i+1}$ 

\quad\quad store transition $(s_i, a_i, r_i, s_{i+1})$ into replay buffer

\quad \textbf{end for} 

\quad randomly sample N transitions from experience replay buffer

\quad \textbf{for} j=1 to N by step k: 

\quad\quad get $n_j$ samples by order 

\quad\quad optimize actor network by minimizing: $\mathcal{L}_{Q} = \frac{1}{n_j} \sum_i(Q_{\theta_{C}}(s_i, a_i))$

\quad\quad optimize critic network by minimizing: $\mathcal{L}_{critic} =\frac{1}{n_j} \sum_i|(Q_{\theta_{C}}(s_i, a_i)-r_i)|$

\quad\quad $M=M+1$

\quad\quad\textbf{if} episode $>$ sharing-start-episode and $M = T$:

\quad\quad\quad get $logits_{A}$ by 
$\pi_{\theta_{A}}(s_i)$

\quad\quad\quad for k in other agents:

\quad\quad\quad\quad get $logits_{k}$ by 
$\pi_{\theta_{k}}(s_i)$

\quad\quad\quad compute $\mathcal{L}_{share} =\frac{1}{n_j} \sum_i\sum_k (\mathcal{L}_{function}(logits_{A},logits_{k})) / (I-1)$
 
\quad\quad\quad where $\mathcal{L}_{function}$ := $\mathcal{L}_{MSE}$

\quad\quad\quad optimize actor network by minimizing $\mathcal{L}_{share}$

\quad\quad\quad $M = 0$

\quad\textbf{end for}

\textbf{end for}
\end{algorithm}

\section{Experiments and Analysis}
\textbf{Multi-agent particle environment.} We perform experiments in multi-agent particle environment (MPE) \cite{lowe2017multi}, which is a simulation environment for multi-agent cooperation and competition, to validate the performance of our method. There are many predefined scenarios, representing some typical cases of collaboration and competition, and allow for modification of existing scenarios. We tested KnowSR in cooperation scenarios based on the MADDPG. In our experiments, KnowSR had a positive impact on accelerating training, primarily by comparing the results of basic algorithms and KnowSR. All the contrasts were based on the same experimental conditions.

\textbf{Simple\_spread scenario.} In the experiments, we take $simple\_spread$ $scenario$ as the testing scenario. In this scenario, agents must cooperate to reach a set of landmarks (targets) without communication. The targets are all the same, and agents are rewarded with  the shared rewards by arriving at all targets as quickly as possible. Meanwhile, agents are also penalized for collision with each other. Each agent can observe agents’ and targets’ positions and agents’ velocity. At the beginning of each episode, the positions of agents and landmarks are reset randomly.

\begin{figure}
\centering
\subfigure[Task I containing 6 agents and 6 targets.]{\label{task1}
\includegraphics[width=0.3\textwidth]{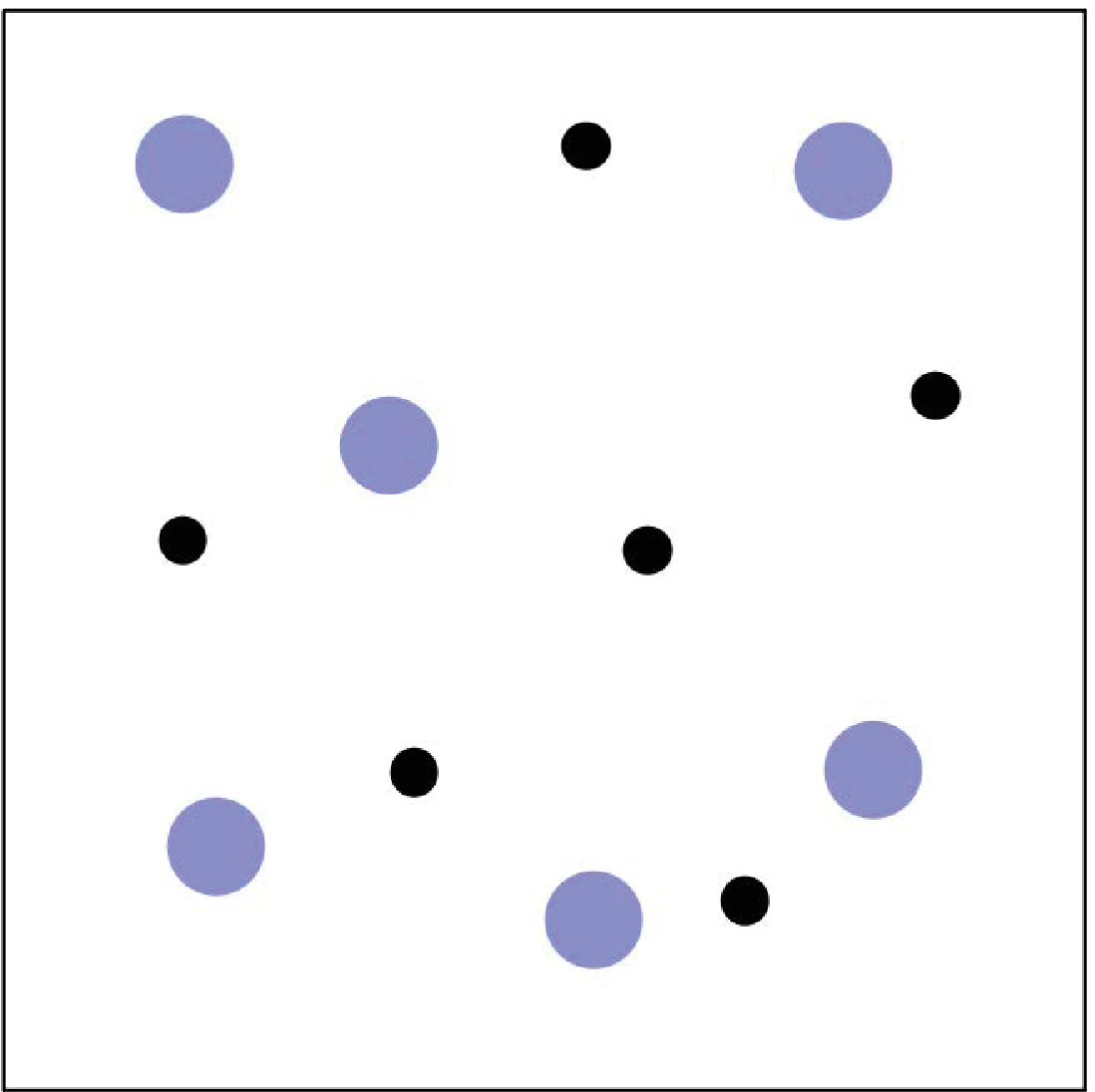}}
\hspace{0.01\linewidth}
\subfigure[Tasks II containing 8 agents and 8 targets.]{\label{task2}
\includegraphics[width=0.3\textwidth]{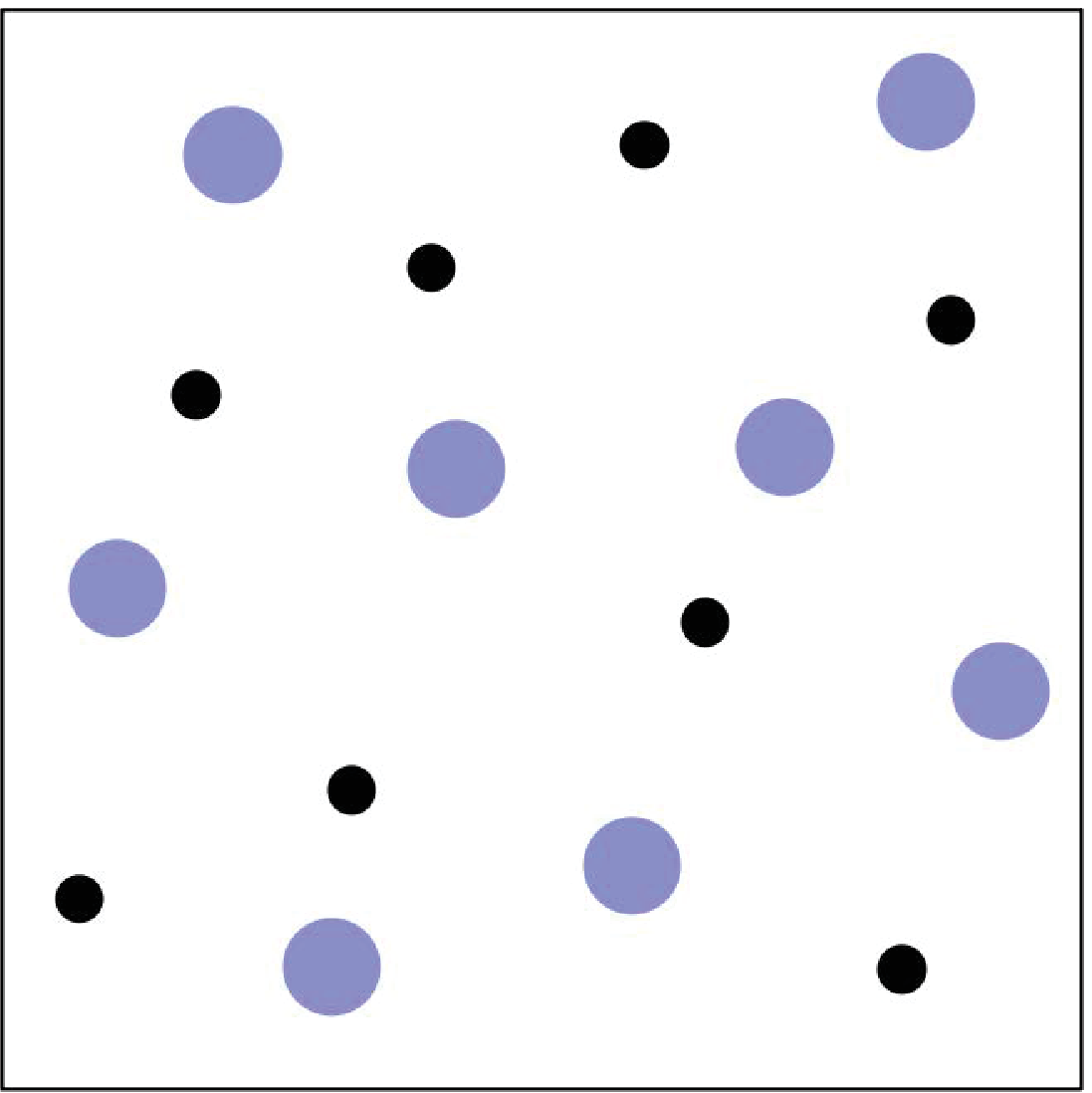}}
\caption{The tasks in Simple\_spread scenario.}
\label{spread}
\end{figure}

\textbf{Experimental settings.} In our experimental settings, as shown in Figure~\ref{spread}, we took the task containing six agents and six targets and as the task containing eight agents and eight targets as the training scenarios, called task I and task II, respectively.  Here, we took the average step reward in each episode to measure the effect of training. The higher the reward, the better was the reward. The actor and critic networks in the experiments were all randomly parameterized by a four-layer fully connected MLP with 64 units per layer. We set 4000 episodes to ensure convergence, every episode consisted of 25 steps, and the networks were updated every four episodes. In simple\_spread, we used MADDPG as the baseline algorithm and implemented KnowSR based on it. All the results and 95\% confidence intervals (95\% CI) are also illustrated in Figures. The red dotted lines represent the best performances in these scenarios. The gray dotted lines denotes the time when agents begin to share knowledge with others. In addition, we added control groups which changed the ratio between self-training times and knowledge sharing times to show how knowledge sharing affect training.

\textbf{A. Task I Experiments}

\begin{figure}[h]
  \centering
  \includegraphics[width=0.6\textwidth]{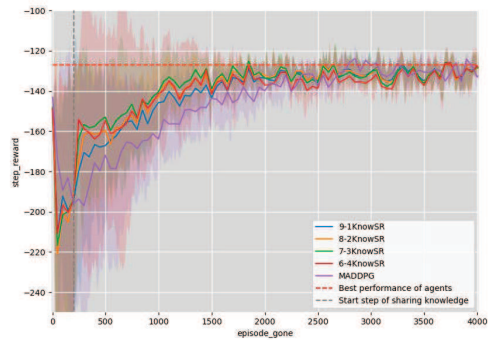}
  \caption{The ratios between the self-training times and the knowledge sharing times was changed at 9:1,8:2,7:3,6:4, namely 9-1KnowSR, 8-2KnowSR, 7-3KnowSR, 6-4KnowSR. This figure clarifies the tendency  in agents performance as the ratio changes in task I.}
  \label{66789}
\end{figure}

In simple\_spread scenario which contains six agents and six targets, we tested several combination of two factors. As shown in Figure~\ref{66789}, we vary the ratio between the self-training times and the knowledge sharing times at 9:1,8:2,7:3,6:4, called 9-1KnowSR, 8-2KnowSR, 7-3KnowSR, 6-4KnowSR. As we mentioned before, the gray line denotes the time that agents begin to share knowledge and the time is set to 200 episodes. Before the time, agents was also trained with MADDPG algorithm.

From Figure~\ref{66789}, compared to MADDPG, it is obvious that the performance of KnowSR in all experiments has been greatly improved at the beginning of sharing knowledge, and the episodes required for training convergence have also been significantly reduced, demonstrating that KnowSR successfully help agents to share knowledge. For example, the agents in 7-3KnowSR firstly show best performance at about 1700 episodes, instead, the agents in MADDPG firstly show best performance at about 2700 episodes. The reason why KnowSR show great performance when agents begin to share knowledge is that the agents have acquired some knowledge which could be regarded as common sense like approaching landmarks and avoid collision at the first 200 episodes and successfully share them in the way of asking for advice. It's worth mentioning that sharing knowledge might be difficult in other formal methods because the knowledge in RL can not always be easily described as formalized expression. Also, we could find that even the self-training times decreased, the more they share the better, which indicates that agents' may learned different knowledge by exploring different subspaces and KnowSR successfully combine their knowledge to make multi-agent system work better.

\begin{figure}[h]
  \centering
  \includegraphics[width=0.6\textwidth]{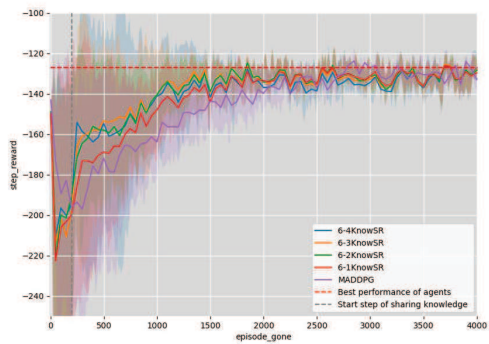}
  \caption{The ratios between the self-training times and the knowledge sharing times was changed at 6:1,6:2,6:3,6:4, namely 6-1KnowSR, 6-2KnowSR, 6-3KnowSR, 6-4KnowSR. This figure indicates how knowledge sharing affects training in task I.}
  \label{66}
\end{figure}

To find out how the frequency of knowledge sharing affects training, we also added control groups that only changed the frequency of knowledge sharing. In addition to the conclusions mentioned above, as illustrated in Figure~\ref{66}, even the self-training times are all the same, proper knowledge sharing times could make better performance. And, it indicates that when agents share knowledge more than once, knowledge can be almost completely shared among agents. For this reason, even the number of sharing knowledge times increase the performance did not get better.

\textbf{B. Task II Experiments}

\begin{figure}
\centering
\subfigure[The ratios between the self-training times and the knowledge sharing times was changed at 9:1,8:2,7:3,6:4, namely 9-1KnowSR, 8-2KnowSR, 7-3KnowSR, 6-4KnowSR in task II.]{\label{86789}
\includegraphics[width=0.475\textwidth]{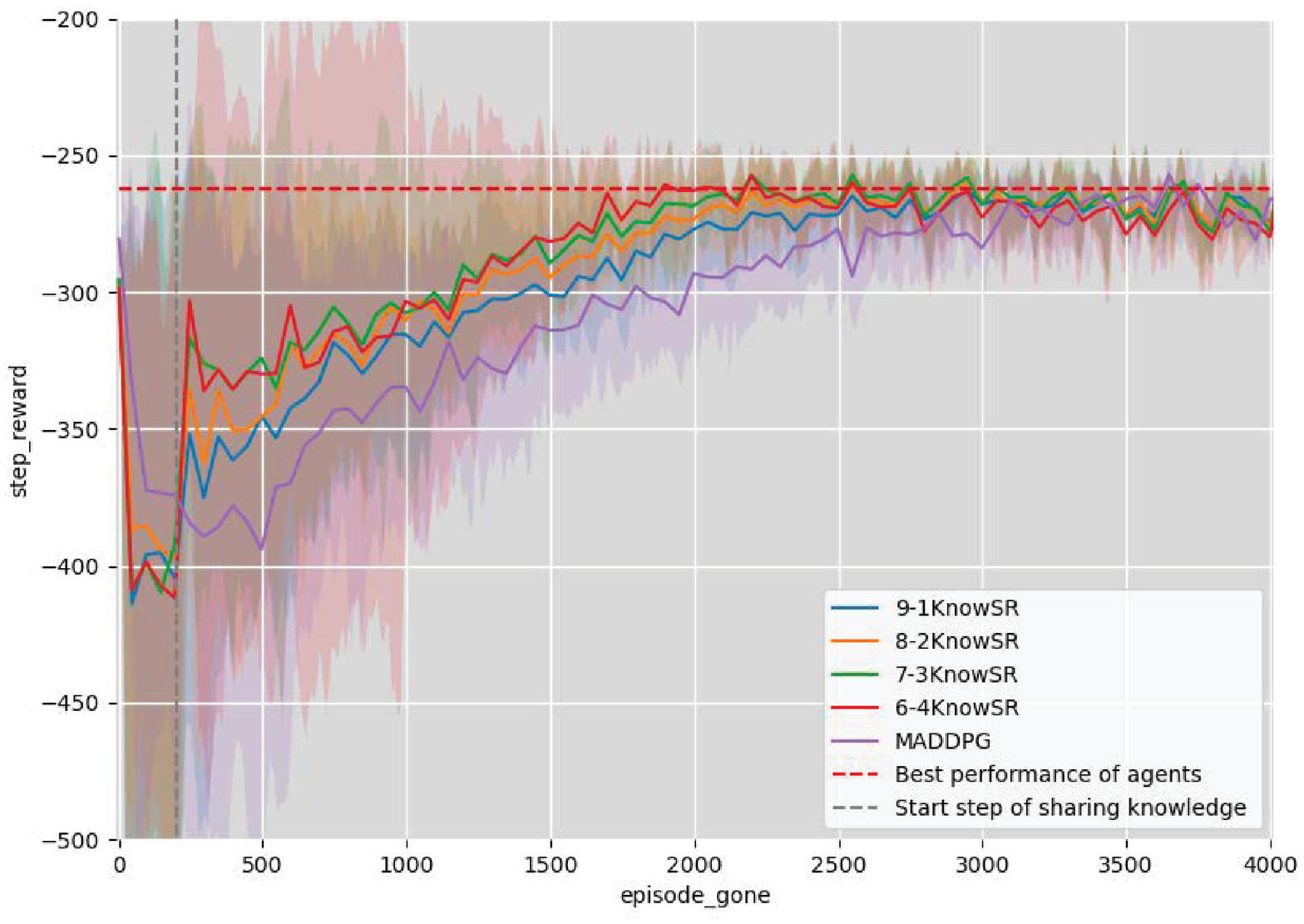}}
\hspace{0.01\linewidth}
\subfigure[The ratios between the self-training times and the knowledge sharing times was changed at 6:1,6:2,6:3,6:4, namely 6-1KnowSR, 6-2KnowSR, 6-3KnowSR, 6-4KnowSR in task II.]{\label{86}
\includegraphics[width=0.475\textwidth]{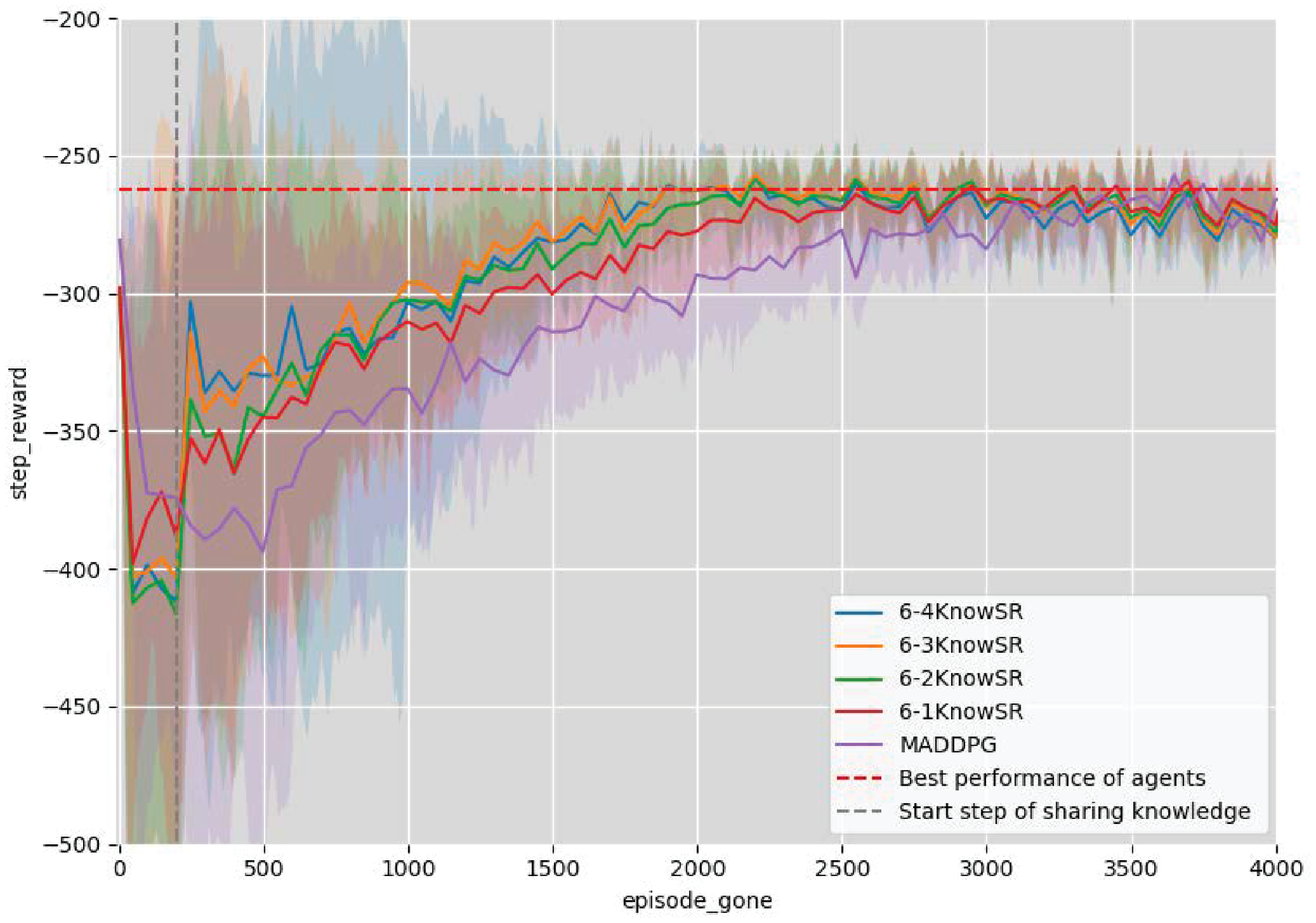}}
\caption{The agents' performance in task II.}
\label{8}
\end{figure}

As shown in Figure~\ref{8}, in simple spread scenario which contains eight agents and eight targets, we also tested several combination of two factors. As we concluded before, the results in Figure~\ref{86789} have also certified that KnowSR successfully help agents to share knowledge with each other, the convergence time is obviously reduced and in the process the rewards is also much higher. According to the results of Figure~\ref{66} and the results in other experimental settings, we didn't see any negative effects of the number of knowledge sharing times except for longer training time.

To explore the impact of KnowSR on the training performance, we listed the average rewards and the time when agents firstly achieve optimum performance of each experimental setting in task II in Table~\ref{table}. It's significant that KnowSR of different combinations of the components all show better performance than MADDPG in not only average reward but also the time to firstly achieve best performance, demonstrating effectiveness of KnowSR’s two-phrase design. We believe that from experiments KnowSR has shown great potential to help multi-agent algorithms to shorten training time and improve training effectiveness. In our opinion, the reason why KnowSR works well is that the solution space can be explored more efficiently and redundant exploration can be avoided by combining knowledge.

\begin{table}
\begin{center}
\caption{Average reward and the time when agents firstly achieve best performance of experiments in task II.}\label{table}
\begin{tabular}{|c|c|c|}
\hline
\diagbox{Method}{Performance}&  Average reward &  Time  \\
\hline
MADDPG& -318.640 & 3692  \\
\hline
9-1KnowSR& -301.776 & 3325 \\
\hline
8-2KnowSR& -294.423 & 2203 \\
\hline
7-3KnowSR& -288.823 & 2187 \\
\hline
6-1KnowSR& -299.223 & 2936 \\
\hline
6-2KnowSR& -292.436 & 2218 \\
\hline
6-3KnowSR& -287.305 & 1849 \\
\hline
6-4KnowSR& -289.049 & 1864 \\
\hline
\end{tabular}
\end{center}
\end{table}

\section{Conclusion}
Knowledge sharing among homogeneous agents is of great practical importance in the MARL fields; however, it is notoriously difficult to express knowledge in formal methods. In this study, we explored a novel knowledge sharing approach for MARL and addressed its accompanying unique challenges, leveraging the KD paradigm. To empirically demonstrate the robustness and effectiveness of KnowSR, we performed extensive experiments on state-of-the-art MARL algorithms in collaborative scenarios. The results prove the effectiveness and robustness of KnowSR under different experimental settings. In our future work, we will extend the diversity of scenarios and tasks, and explore more efficient knowledge sharing methods.

\bibliographystyle{splncs04} 

\end{document}